\documentclass[10pt]{article}
\usepackage[numbers,sort&compress]{natbib}

\usepackage{graphicx}
\usepackage{subfigure}
\usepackage{amsfonts, amsmath}
\usepackage{authblk}
\usepackage{hyperref}

\usepackage{booktabs}
\newcommand{\figureref}[1]{Figure~\ref{#1}}
\newcommand{\tableref}[1]{Table~\ref{#1}}
\usepackage{xspace}
\newcommand{\R}{\mathbb{R}}
\newcommand{\bob}{\textsc{rNCA}\xspace}
\newcommand{\bobs}{\textsc{rNCA's}\xspace}
\usepackage[T1]{fontenc}

\begin{document}
\title{\bob: Self-Repairing Segmentation Masks}
\author[1,*]{Malte Silbernagel}
\author[1,*]{Albert Alonso}
\author[1,2]{Jens Petersen}
\author[1]{Bulat Ibragimov}
\author[1,3]{Marleen de Bruijne}
\author[1]{Madeleine K.\ Wyburd}

\affil[1]{Department of Computer Science, University of Copenhagen, Copenhagen, Denmark}
\affil[2]{Department of Oncology, Rigshospitalet, Copenhagen, Denmark}
\affil[3]{Department of Radiology and Nuclear Medicine, Erasmus MC -- University Medical Center Rotterdam, The Netherlands}
\affil[*]{Contributed equally}

\date{}
\maketitle
\begin{abstract}
Accurately predicting topologically correct masks remains a difficult task for general segmentation models, which often produce fragmented or disconnected outputs.
Fixing these artifacts typically requires handcrafted refinement rules or architectures specialized to a particular task. 
Here, we show that Neural Cellular Automata (NCA) can be directly repurposed as an effective refinement mechanism, using local, iterative updates guided by image context to repair segmentation masks.
By training on imperfect masks and ground truths, the automaton learns the structural properties of the target shape while relying solely on local information.
When applied to coarse, globally predicted masks, the learned dynamics progressively reconnect broken regions, prune loose fragments and converge towards stable, topologically consistent results.
We show how refinement NCA (\bob) can be easily applied to repair common topological errors produced by different base segmentation models and tasks:
for fragmented retinal vessels, it yields 2--3\% gains in Dice/clDice and improves Betti Errors, reducing $\beta_0$ errors by 60\% and $\beta_1$ by 20\%; for myocardium, it repairs 61.5\% of broken cases in a zero-shot setting while lowering ASSD and HD by 19\% and 16\%, respectively.
This showcases NCA as effective and broadly applicable refiners.
Code available at \url{www.github.com/maltesilber/rnca}.
\end{abstract}
\newpage

\begin{figure}[b]
\includegraphics[width=\linewidth, trim=15 25 25 15, clip]{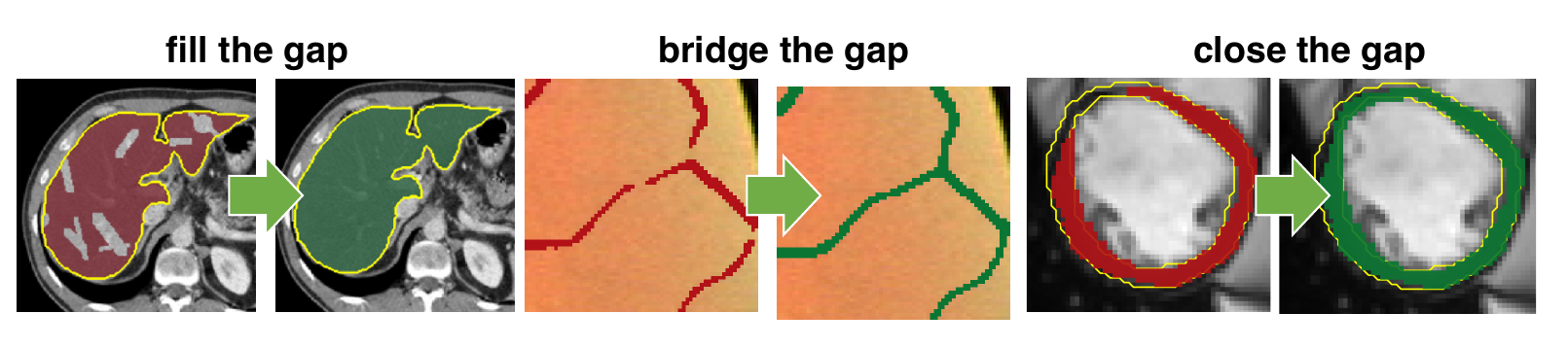}
\caption{\textbf{Examples of topologies repairs achieved by \bob}, including filling, reconnection and closing.}
\label{fig:showcase}
\end{figure}

\section{Introduction}
\label{sec:intro}

Modern segmentation networks achieve strong pixel- and voxel-level accuracy across medical tasks, but often fail to preserve topological correctness when dealing with thin branches, closures, and continuity~\cite{Long_CVPR_2015_FCN,Ronneberger_MICCAI_2015_UNet,Huang_ICCV_2019_CCNet,Yuan_ECCV_2020_OCR,Liu_ICCV_2021_Swin}.
This results in fragmented masks with gaps degrading their reliability for quantitative downstream image processing tasks, critical for many clinical applications~\cite{wyburd2024anatomically}.

To mitigate such failures, many works modify the segmentation model and the training pipelines themselves through, for example,  topology-aware losses, diffeomorphic transformations or surrogate representations or by focusing the architecture in a specific anatomical prior~\cite{Clough_TPAMI_2020_TopoLossPH,shit2021cldice,Kervadec_MedIA_2021_BoundaryLoss, wyburd2021teds, alonso2023fast}.
Of particular interest to the medical imaging community is the detection of thin tubular structures, where anatomically plausible solutions are paramount and challenging to enforce in generic models~\cite{kirchhoff2024skeleton, huang2024meta, song2025optimized, amiri2024centerline}.
While specialized methods improve structural fidelity, their strong ties to a particular problem limit their applicability across domains and make them harder to translate to new settings.

Because of this, many approaches leave the general segmentation model unchanged and use post-processing steps to improve output coherence. Classical post-processing methods promote local smoothness but struggle to repair complex topology~\cite{Boykov_ICCV_2001_GraphCuts, Soille_2003_MorphologyBook}, while learned refiners, such as CRF-RNN~\cite{Zheng_ICCV_2015_CRFRNN}, integrate similar operations into an end-to-end framework, yet remain limited to local consistency rather than topological repair. More recent work embeds topological priors directly into the refinement step~\cite{zdyb2025spline, dima2025parametric}, although these methods remain tailored to specific problem settings.

In contrast to these,  we treat mask repair as a problem-agnostic \emph{learned local dynamical process} applied after segmentation.
Neural Cellular Automata (NCA)~\cite{mordvintsev2020growing} provide a natural framework for learning local update rules that iteratively evolve spatial states, and whose dynamics can be externally conditioned~\cite{sudhakaran2022goal}.
Prior work has already shown that NCA can generate coherent end-to-end segmentations in both 2D and 3D~\cite{kalkhof2023med,kalkhof2023m3d,ranem2025ncadapt} for low compute solutions but often at the cost of reduced accuracy.
Here, we use NCA as a plug-in refiner, referred to as \bob, that learns iterative growth and pruning dynamics to repair initial segmentation masks.
The method operates without retraining the base model and generalizes across segmentation architectures, yielding more connected and more structurally and anatomically plausible masks (\figureref{fig:showcase}).

\section{Related Work}
\label{sec:related}

\paragraph{Neural Cellular Automata}
NCA extend classical Cellular Automata by making the local update rule learnable~\cite{automata,mordvintsev2020growing}, enabling stable, self-organizing structures through iterative local updates.
Most NCA work focuses on pattern growth, external conditioning, or spatio-temporal behaviors~\cite{pajouheshgar2024noisenca,sudhakaran2022goal,pajouheshgar_DyNCARealTimeDynamic_2023}, typically for a single target shape or synthetic tasks.
Recent extensions explore probabilistic variants~\cite{palm2022variational,faldor2024cax,kalkhof2025parameter} and lightweight biomedical applications~\cite{kalkhof2023med,kalkhof2023m3d,ranem2024nca}, though adoption in medical imaging remains limited. To the best of our knowledge, this is the first work to propose NCA as a refinement tool, potentially opening new directions for future research.

\paragraph{Segmentation Refinement}
Refinement has emerged as a practical alternative to modifying or retraining large segmentation models, allowing targeted correction of their systematic errors.
Classical post-processing methods such as morphological filling and pruning~\cite{Soille_2003_MorphologyBook} address only simple artifacts, whereas model-tied refiners (e.g., PointRend~\cite{Kirillov_CVPR_2020_PointRend}, RefineMask~\cite{zhang2021refinemask}, Deep Closing~\cite{wu2024deep}) attach task-specific heads which require retraining.
On the other hand, model-agnostic approaches act on noisy masks using boundary or global information ~\cite{tang2021look,yuan2020segfix,zhou2020deepstrip,cheng2020cascadepsp, SegRefiner} to further finetune segmentation in a single forward pass.
An interesting perspective is that of iterative strategies~\cite{lagergren2020region,lagergren2023few}, which show that repeated local updates can successfully stabilize topology and promote solutions that promote mask connectivity.
Our work builds on the view that segmentation refinement is essentially the problem of learning local repair dynamics. 
NCA already operates through learned local updates that evolve a state toward stable shapes, which aligns directly with how topological corrections must be applied.

\begin{figure}[tb]\label{fig:overview}
\includegraphics[width=\textwidth]{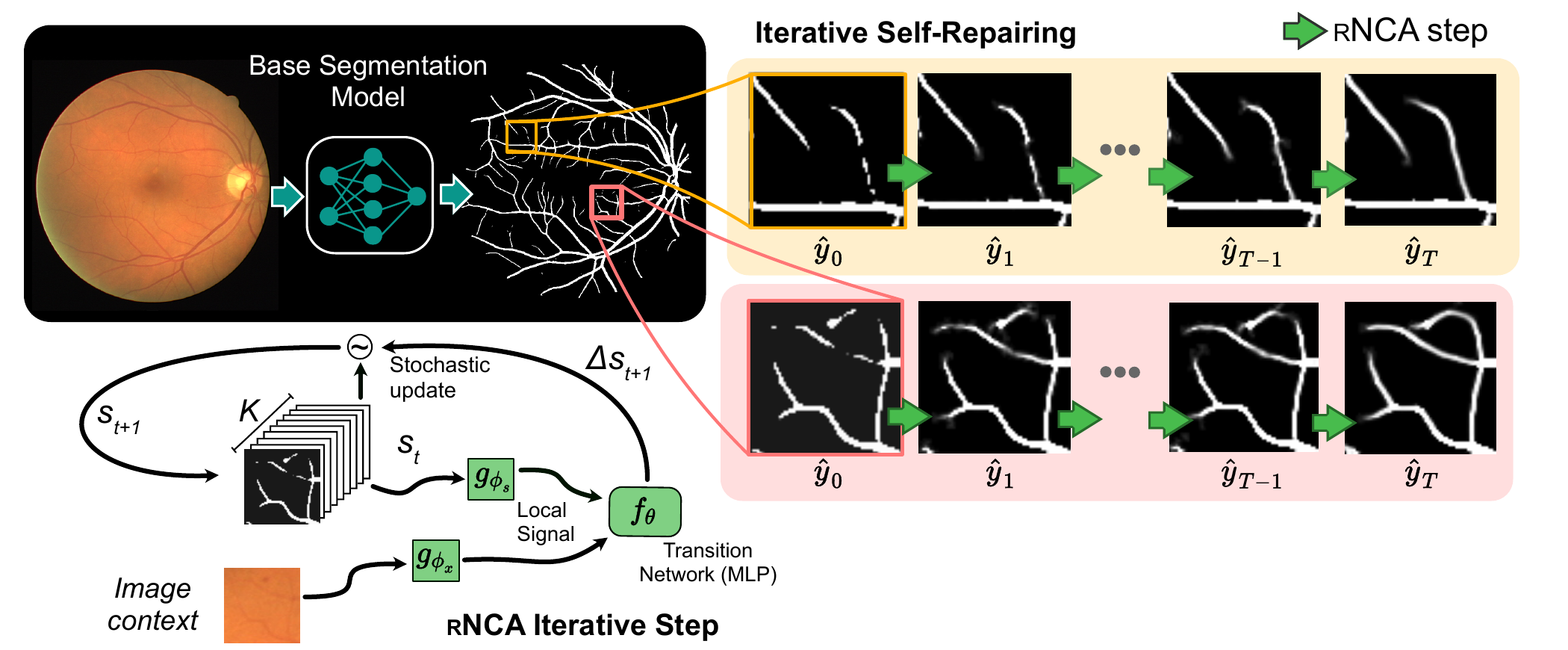}
\caption{\textbf{Overview of \bob refinement pipeline}. A coarse mask is refined through iterative local updates guided by image context.}
\label{fig:diagram}
\end{figure}

\section{Method}
\label{sec:method}
We model segmentation refinement as an iterative local state evolution inspired by the NCA~\cite{mordvintsev2020growing}.
Given an input image $x$ and an initial mask $\hat{y}_0$, the refiner maintains a per-pixel state vector $s_t$ with $K$ channels. The first channel of $s_t$ encodes the observable mask $\hat{y_t}$ and the remaining $K-1$ channels capture the internal dynamics of the system of timestep $t$. An overview is shown in \figureref{fig:overview}.

\subsection{Design philosophy}
\label{subsec:design}
The model does not encode any task-specific topology.
Instead, it learns generic repair dynamics based on the perturbative states seen during trainings.
This keeps the refiner architecture-agnostic, dataset-agnostic and simple to integrate into existing pipelines.
We keep the refiner tightly close to the original NCA formulation~\cite{mordvintsev2020growing} to showcase that the benefits presented here are not due to specializing it to our problem, but it performs as intended without bespoken methods.

\subsection{Local perception and update rule}
\label{subsec:nca}
At each iteration, every pixel updates its state using only its local neighborhood from the current state $s_t$ and the input image $x$. To extract these local features, we use a learnable perception operator $g_\phi$ composed of two $3{\times}3$ convolutional layers for $s_t$ and $x$, formally:
\begin{equation}
    g_\phi(s_t, x) = \big[\, g_{\phi_s}(s_t) \; \| \; g_{\phi_x}(x) \,\big],
\end{equation}
where $g_{\phi_s}(s_t)$ and $g_{\phi_x}(x)$ denote the respective perceive networks and $\; \| \; $ denotes the channel-wise concatenation. 
The output of the percieve network $g_\phi$ are passed to a small learned transition function $f_\theta$, defined as a multilayer perception (MLP), which results in the update:
\begin{equation}
    s_{t+1} = s_t + f_\theta(g_\phi(s_t, x)).
\end{equation}
This formulation enforces strictly local reasoning for state dynamics while allowing information to propagate globally through iterative updates.

Next, masking of \textit{alive pixels} is applied so that only pixels within or near the active regions of $s_t$ and $s_{t+1}$ are allowed to modify the state, leading to growth-pruning dynamics that fix errors while preserving current structures. This contrasts with standard segmentation networks, where each pixel can freely update its predicted value. 

\subsection{Training repair dynamics}
\label{subsec:training}
The refiner is indirectly trained to repair topological errors by learning dynamics that reconstruct initial perturbed masks, which are either generated by an imperfect segmentation method or by manually corrupting ground-truths (e.g breaking topology and connectivity).
This aims to emulate systematic segmentation artifacts.
Training uses NCA trajectories and employs classic NCA techniques by using a sample pool strategy that continuously samples and updates initial states, forcing the cellular automaton to learn both how to grow the target pattern from a seed and how to maintain it once formed. This effectively creates the target pattern as an attractor in the system's dynamics.
To do so, the loss is computed at a randomly sampled late step $t^* \sim U(T/2, T)$ using the standard MSE loss between the ground truth $y$ and the prediction at step $t^*$:
\begin{equation}
\mathcal{L}~=~|\hat{y}_{t^*} - y\|_2^2.
\end{equation}
For a more detailed explanation on the implementation see in the Appendix ~\ref{app:model}.

\subsection{Inference}
\label{subsec:inference}
At test time, the NCA is used as a plug-in refinement operator.
Given any predicted mask $\hat{y}_0$ from an arbitrary segmentation model, we initialize latent channels and run a fixed number of update steps to produce a refined mask $\hat{y}_T$.
Since the update rule is fully local, performance is independent of mask size and thus can be used in input images of arbitrary resolution.

\section{Experiments}
\label{sec:exp}
We evaluate \bob across three representative classes of topological artifacts of general segmentation methods, as shown in \figureref{fig:showcase}.

First, we reconstruct liver volumes using segmentation with synthetically-generated gaps (\ref{subsec:exp-fill}). Then, we explore reconnecting thin structures (\ref{subsec:exp-bridge}) exemplified by retinal vessel segmentation masks, while also showing how the method performs on fixing broken rings as can occur in cardiac segmentation in a zero-shot scenario (\ref{subsec:exp-close}).
Across these tasks, \bob shares the same architecture and training protocol, with the only difference being the data used to train them, see Appendix \ref{app:model}. Parameters were chosen empirically, ablation shown in Appendix \ref{app:drive}.

To quantify the performance of the methods, we report overlap metrics; Dice and centerline Dice (clDice), together with geometric errors such as average symmetric surface distance (ASSD) and Hausdorff distance (HD), and Betti-number deviations as a proxy for topological correctness \cite{millson1976first}. Information on the datasets and baseline implementations are available for each experiment in the Appendix \ref{app:liver-metrics}-\ref{app:hearts}.

\subsection{Fill the Gaps}
\label{subsec:exp-fill}

\begin{figure}[tb]
\includegraphics[width=\textwidth]{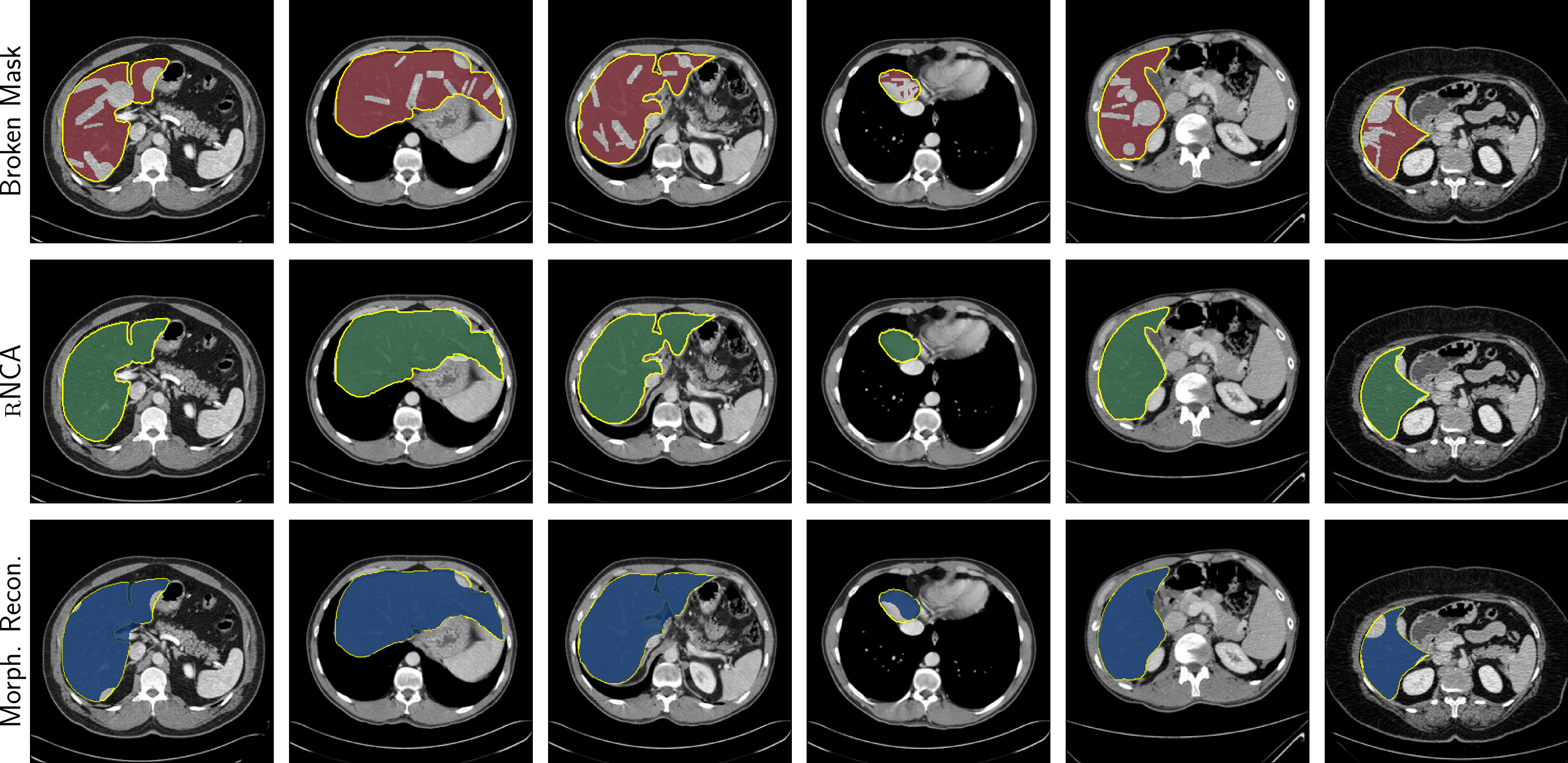}
\caption{\textbf{Liver mask restoration}. Input slice with corrupted mask (top), \bob output (middle), and morphological reconstruction (bottom), with ground-truth contours in yellow.}
\label{fig:fill}
\end{figure}
Holes and missing patches are common artifacts in segmentations of genus-0 structures, where local inconsistencies in pixel-wise classification may lead to anatomically implausible masks.
To correct for those, classical morphological operations, such as filling~\cite{Soille_2003_MorphologyBook}, are commonly applied and, while effective and simple, become problematic when dealing with objects of non-uniform outline and large degrees of corruptness.

To assess whether \bob can repair such defects, we construct a synthetic variant of the CHAOS liver dataset~\cite{CHAOSdata2019} by removing regions of varying size and shape from ground-truth masks.
The cohort includes 20 CT patients, which we split into training/validation/test sets at the patient level, and from which we create multiple perturbed states to diversion examples of incomplete masks, see Appendix \ref{app:liver-metrics}.

As seen in \figureref{fig:fill}, \bob restores missing regions while preserving the underlying object geometry. Both morphological operations and \bob improve the Dice score from $0.80$ to $0.96$ and $0.98$, respectively. However, as shown in \figureref{fig:fill}, the morphological operations often over-segment the liver in areas of tight folds or convoluted areas, whereas \bob learns the anatomical shape, preserving these features. This result is reflected in the distance measures, where \bob improves ASSD from $0.67$ to $0.08$ and HD decreases from $10.5$ to $4.3$, a $58\%$ and $43\%$ improvement compared to morphological operations.

This showcases the benefit of \bob on situations where context and structural traits can be exploited for improved refinement.
Full metrics appear in Appendix~\ref{app:liver-metrics}.

\begin{figure}[tb]
\begin{minipage}[c]{0.6\linewidth}
  \centering
  {\includegraphics[width=\textwidth]{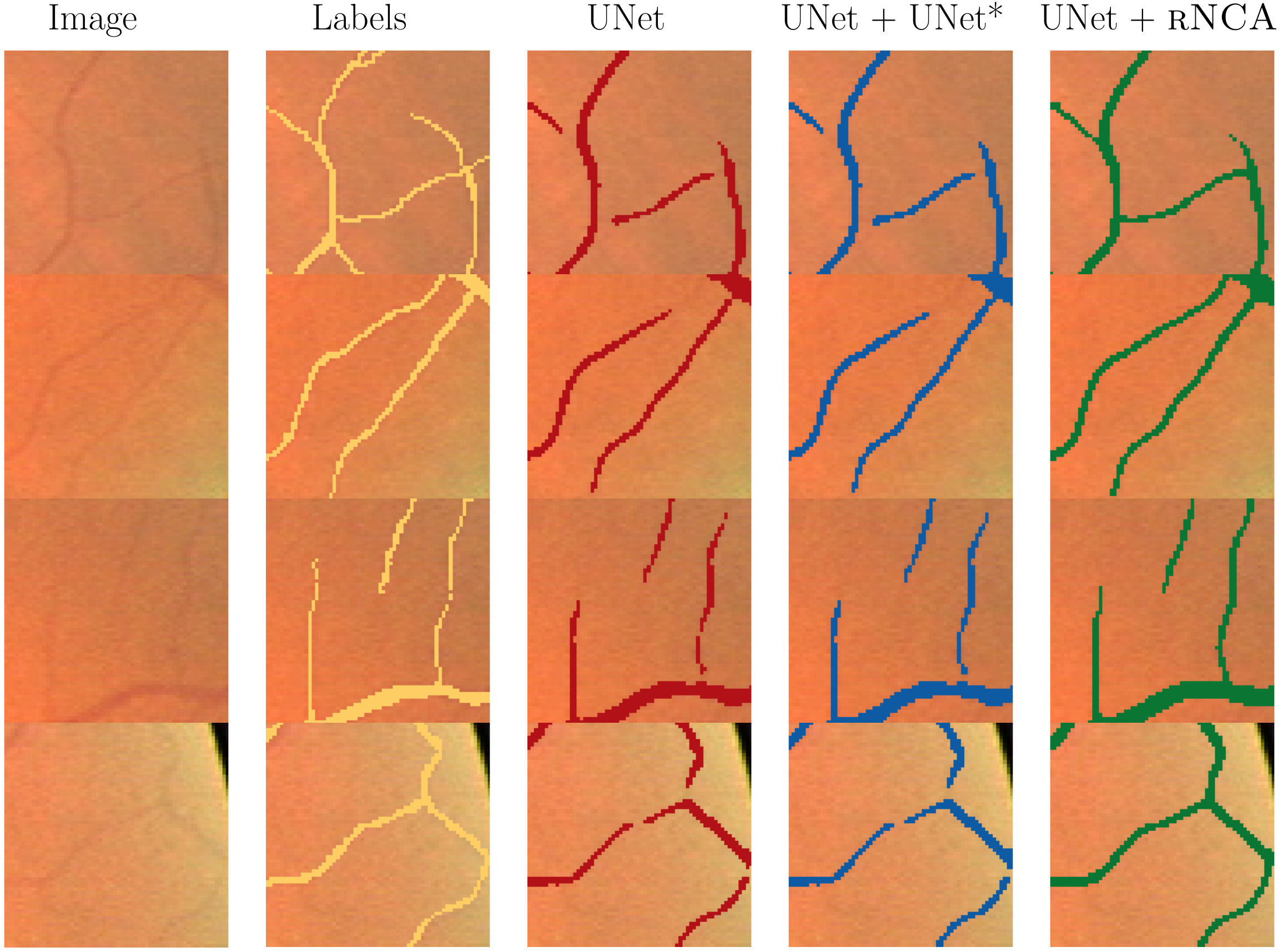}}
\end{minipage}
\begin{minipage}[c]{.375\linewidth}
 \centering
    \includegraphics[width=\linewidth]{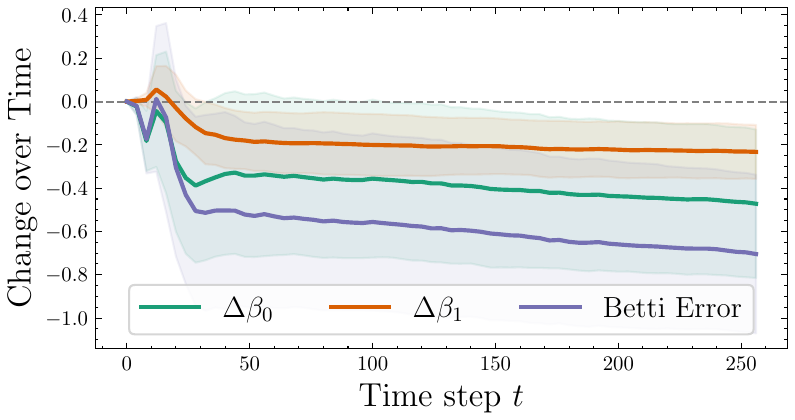} \\[6pt]
    \includegraphics[width=0.9\linewidth]{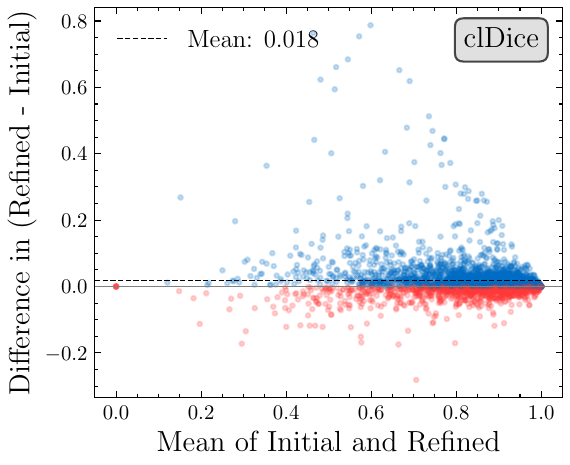}
\end{minipage}
\caption{\textbf{Retina vessel refinement.} (left) UNet predictions masks with UNet and \bob refinements. (right, upper) $\beta_0$, $\beta_1$ and Betti error progression during the refinement steps. Here we display the mean and std over all images. (right, lower) Displays a Bland-Altman Plot comparing initial and refined clDice scores of each patch. Plots for Dice and clDice are shown in Appendix \ref{app:drive}.}
\label{fig:bridge}
\end{figure}

\subsection{Bridge the Gaps}
\label{subsec:exp-bridge}
Thin tubular structures are among the most error-prone targets for general segmentation models, as pixel-wise approaches often produce fragment masks at high uncertainty regions.
To examine whether \bob can repair broken connectivity, we explore retinal vessel segmentation on the DRIVE dataset~\cite{staal2004ridge}.
Here, unlike in the liver task, broken segments and discontinuities are observed in the UNet segmentation of both healthy and pathological retinal vessels.
Therefore, we train \bob directly on those to expose it to realistic break patterns instead of synthetic perturbations.
Evaluation is performed on predictions from both UNet used during training of \bob and unseen Swin model to assess cross-architecture generalization.

\figureref{fig:bridge} qualitatively demonstrates that \bob effectively bridges gaps in fragmented vein segmentations, and that the biggest improvement occurs in the initial steps and maintains a steady state from there. %
It also compares initial and refined clDice scores of individual patches, and suggests that, on average \bob provides modest improvements while maintaining stability across all initial scores. 
Consistent with these observations, \tableref{tab:bridge-quantitative} shows that \bob increases both Dice and clDice, and notably reduces Betti-number deviations from the base-method predictions. 
Despite being trained solely on UNet outputs, the refiner generalizes well and similarly improves masks generated by Swin.
To benchmark against prior refinement methods, we compare \bob with SegFix~\cite{yuan2020segfix}, SegRefiner~\cite{SegRefiner}, DenseCRF~\cite{Krahenbuhl_NIPS_2011_DenseCRF}, and a conditional UNet trained to refine the initial guess, all of which are trained on the same dataset with UNet predictions until convergence, see Appendix \ref{app:drive}. 

\bob outperforms all baselines across metrics and base models. In particular, \bob has large improvements in topology performance, showing its ability to indirectly learn the topological features from the training dataset.
We note that, unlike \bobs iterative refinement, these approaches rely on a single-pass correction, which may limit their ability to handle challenging predictions.

\begin{table}[tb]
\label{tab:bridge-quantitative}
\caption{
Comparison of baseline segmentation models with their refined outputs on DRIVE.
Metrics include overlap quality (DICE, clDICE) and topological accuracy ($\beta_0$, $\beta_1$). The best result from each metric is shown in bold.
}
\begin{tabular}{lcccc}
\toprule
\textbf{Model} &
\textbf{DICE} $\uparrow$ &
\textbf{clDICE} $\uparrow$ &
$\Delta\beta_0$ $\downarrow$ &
$\Delta\beta_1$ $\downarrow$ \\
\midrule
\multicolumn{5}{c}{{\bfseries Refiners trained and evaluated on UNet}} \\
\midrule
U\text{-}Net &
$0.761 \pm 0.027$ &
$0.802 \pm 0.028$ &
$0.965 \pm 0.291$ &
$1.366 \pm 0.535$ \\
\, + U\text{-}Net* &
$0.761 \pm 0.027$ &
$0.804 \pm 0.027$ &
$0.917 \pm 0.291$ &
$1.389 \pm 0.535$ \\
\, + SegFix &
$0.759 \pm 0.027$ &
$0.800 \pm 0.029$ &
$0.957 \pm 0.282$ &
$1.372 \pm 0.518$ \\
\, + SegRefiner &
$0.760 \pm 0.026$ &
$0.804 \pm 0.029$ &
$0.941 \pm 0.279$ &
$1.358 \pm 0.524$ \\
\, + DenseCRF &
$0.760 \pm 0.026$ &
$0.803 \pm 0.028$ &
$0.712 \pm 0.230$ &
$1.367 \pm 0.527$ \\
\, + \bob &
$\mathbf{0.777 \pm 0.029}$ &
$\mathbf{0.820 \pm 0.029}$ &
$\mathbf{0.532 \pm 0.193}$ &
$\mathbf{1.094 \pm 0.485}$ \\
\midrule
\multicolumn{5}{c}{{\bfseries Refiners trained on UNet, evaluated on unseen Swin model}} \\
\midrule
Swin &
$0.767 \pm 0.026$ &
$0.804 \pm 0.025$ &
$1.547 \pm 0.393$ &
$1.253 \pm 0.529$ \\
\, + U\text{-}Net* &
$0.767 \pm 0.026$ &
$0.806 \pm 0.025$ &
$1.408 \pm 0.393$ &
$1.291 \pm 0.529$ \\
\, + SegFix &
$0.765 \pm 0.026$ &
$0.802 \pm 0.028$ &
$1.419 \pm 0.352$ &
$1.265 \pm 0.519$ \\
\, + SegRefiner &
$0.766 \pm 0.025$ &
$0.806 \pm 0.026$ &
$1.485 \pm 0.391$ &
$1.255 \pm 0.517$ \\
\, + DenseCRF &
$0.767 \pm 0.026$ &
$0.806 \pm 0.023$ &
$0.902 \pm 0.213$&
$1.259 \pm 0.523$ \\
\, + \bob &
$\mathbf{0.776 \pm 0.032}$ &
$\mathbf{0.816 \pm 0.030}$ &
$\mathbf{0.646 \pm 0.187}$ &
$\mathbf{1.122 \pm 0.475}$ \\
\bottomrule
\end{tabular}
\end{table}

\subsection{Close the Gap}
\label{subsec:exp-close}
Breaks in the myocardial rings are a frequent failure mode in cardiac MRI segmentation and directly affect perimeter-based clinical measurements~\cite{wyburd2024anatomically}.
To study the applicability of \bob on such scenarios where contour completeness is broken, we use \emph{ACDC} myocardium masks~\cite{bernard2018deep} and generate perturbations via thinning, opening, and thickening on the ground truth.
Unlike the retinal vessel tasks, most of the predictions from segmentation models are already topologically correct, so in order to avoid overcorrection from \bob, we include unperturbed ground truths into the dataset, see Appendix \ref{app:hearts} for more details.

\begin{figure}[htb]
\includegraphics[width=.9\linewidth]{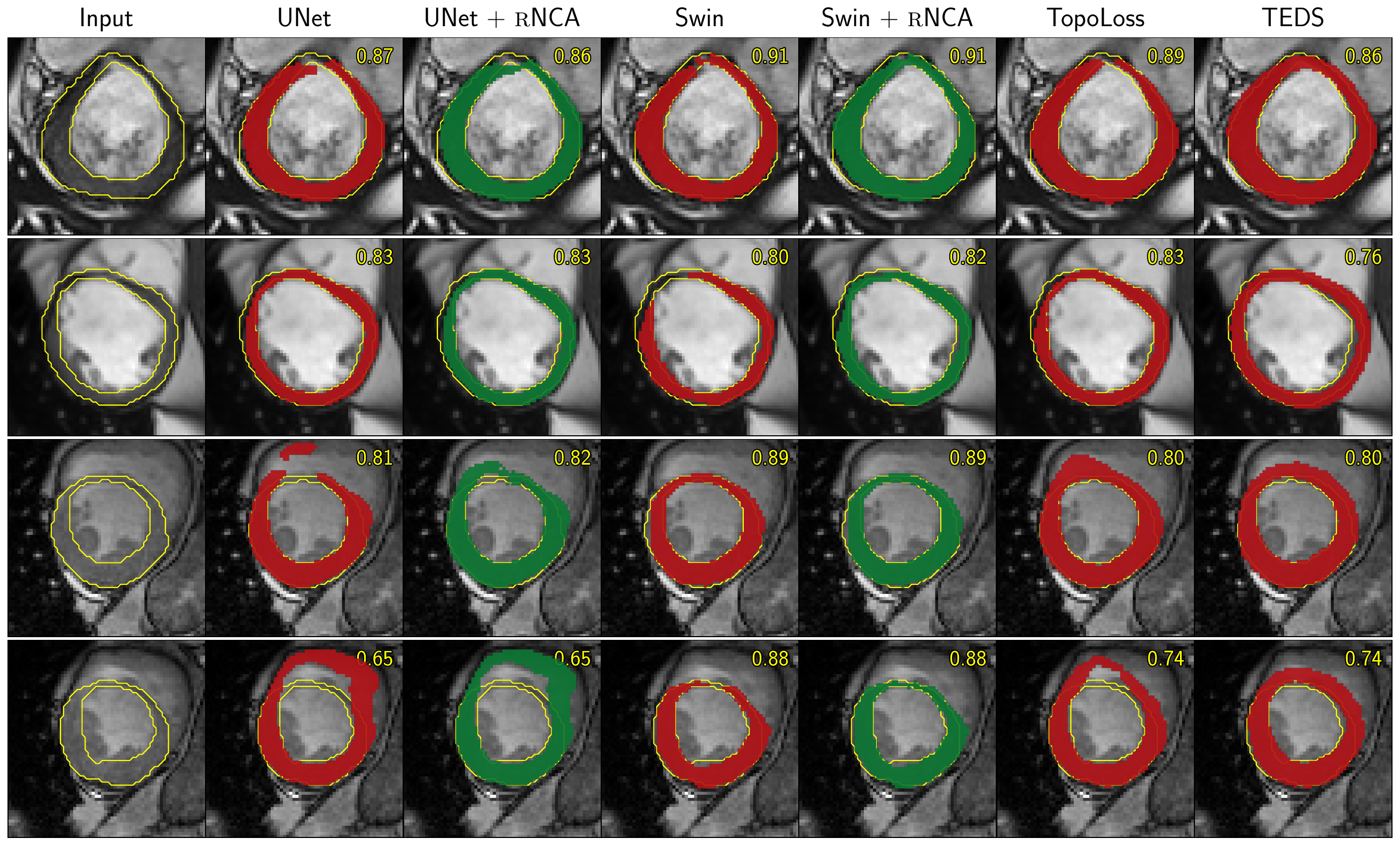}
\caption{\textbf{Myocardium ring correction.} Predictions before and after \bob, alongside TopoLoss and TEDS. Dice scores shown in yellow; bottom row illustrates a failure case (UNet+\bob), with more shown in Appendix \ref{app:hearts}.}
\label{fig:close-qualitative}
\end{figure}

We evaluate \bob on predictions from independently trained UNet and Swin models, each exhibiting distinct topological errors, with implementation described in Appendix \ref{app:hearts}.
Results are then reported both on the full test set in \tableref{tab:close-all} as well as on those cases where the ring topology was not preserved in the initial segmentation~(\tableref{tab:broken-combined}).

Refining both models’ predictions with \bob improves segmentation performance across all metrics, with particularly strong gains on cases that initially had non-closed loops (\tableref{tab:broken-combined}).
We further compare \bobs with two topology-aware approaches: a topological loss function TopoLoss~\cite{hu2019topology} and a topology-preserving deformation model, TEDS-Net~\cite{wyburd2021teds}, explicitly designed for myocardium heart segmentation. 

\figureref{fig:close-qualitative} shows qualitative examples of myocardium repair.
Here, \bob is shown to repair gaps across the networks. 
Although TEDS-Net has 100\% topology accuracy, it often produces overly segmented, misaligned rings (see Appendix \ref{app:hearts}), whereas, \bob is able to join disconnected regions following the real contours.
This figure also highlights a limitation of the method: as \bob is a local refinement process, if an initial mask is particularly poor (bottom row and Appendix \ref{app:hearts}), it can not effectively repair it. 

\begin{table}[htb]
\centering
\label{tab:close-all}
\caption{Performance on all myocardium test slices (100 slices). \bob preserves global segmentation accuracy and Betti-number stability.}
\begin{tabular}{lcccc}
\toprule
\bfseries Model & \bfseries DICE $\uparrow$ & \bfseries ASSD $\downarrow$ & $\Delta\beta_0 \downarrow$ & $\Delta\beta_1 \downarrow$ \\
\midrule
UNet & 
$0.868 \pm 0.155$ &
$0.81 \pm 0.90$ &
$0.04 $ &
$0.07$ \\
\, + \bob &
$0.871 \pm 0.145$ &
$0.83 \pm 1.15$ &
$0.03$ & 
$0.05$ \\
Swin &
$0.855 \pm 0.173$ &
$0.79 \pm 0.72$ &
$0.09$ &
$0.12$ \\
\, + \bob  &
$0.859 \pm 0.166$ &
\textbf{0.75 $\pm$ 0.53} &
$0.02$ &
$0.06$ \\
TopoLoss&
$\mathbf{0.876} \pm \mathbf{0.150}$ &
$0.83 \pm 1.26$ &
$0.02$ & 
$0.06$ \\
TEDS &
$0.848 \pm 0.140$ &
$0.94 \pm 0.75$ &
\textbf{0.00} &
\textbf{0.00 } \\
\bottomrule
\end{tabular}
\end{table}

\begin{table}[htb]
\centering
\label{tab:broken-combined}
\caption{Performance on slices where the predicted myocardium ring is broken. 
The upper block corresponds to UNet failures and the lower block to Swin failures. 
}
\begin{tabular}{lcccc}
\toprule
\bfseries Model & \bfseries DICE $\uparrow$ & \bfseries HD $\downarrow$ & \bfseries ASSD $\downarrow$ & \bfseries Topo. $\uparrow$ \\
\midrule
\multicolumn{5}{c}{\bfseries UNet broken subset (8 slices)} \\
\midrule
UNet        & 0.490 $\pm$ 0.345 & 13.48 $\pm$ 6.30 & 2.96 $\pm$ 2.30 & 0/8  \\
\, + \bob & \textbf{0.517 $\pm$ 0.318} & 12.11 $\pm$ 8.60 & 3.27 $\pm$ 3.34 & 2/8  \\
TopoLoss     & 0.515 $\pm$ 0.336 & 10.47 $\pm$ 6.16 & 3.53 $\pm$  3.34& 3/8 \\
TEDS         & 0.499 $\pm$ 0.293 &  \textbf{8.69 $\pm$ 4.96 } & \textbf{2.57 $\pm$ 1.86} & \textbf{8/8}\\
\midrule
\multicolumn{5}{c}{\bfseries Swin broken subset (13 slices)} \\
\midrule
Swin        & 0.539 $\pm$ 0.329 & 7.09 $\pm$ 4.22 & 2.01 $\pm$ 1.65  & 0/13\\
\, + \bob & 0.569 $\pm$ 0.325 & \textbf{5.98 $\pm$ 4.07} & \textbf{1.63 $\pm$  1.16}& 8/13 \\
TopoLoss    &\textbf{ 0.611 $\pm$ 0.293} & 8.10 $\pm$ 5.79 & 2.65 $\pm$ 2.87 & 7/13 \\
TEDS        & 0.600 $\pm$ 0.264 & 6.91 $\pm$ 4.57 & 2.08  $\pm$ 1.61 & \textbf{13/13} \\
\bottomrule
\end{tabular}
\end{table}

\section{Discussion}
\label{sec:discussion}
Across three structurally distinct repair tasks, \bob has consistently improved topological correctness while minimally affecting already good predictions.
We have seen that training both on corrupted labels and on model predictions yields effective repair dynamics that generalize across architectures.
We have also shown that, due to its simplicity, the method is easy to port to different datasets without having to fine tune its hyperparameters.
These results showcase that local refinement is sufficient to correct a broad range of structural failures present in popular segmentation models.

The experiments also show why NCA naturally fits the refinement task.
Corrections are applied only where the state dynamics detect inconsistencies, much similar to biological systems repair mechanisms, while limiting the risk of over-refinement.
This trait is particularly relevant in the setting of medical applications, where excessive and anisotropic mask postprocessing can produce large mask alterations that are often undesirable.

Nevertheless, the focus on local dynamics also presents its limitations.
For too broken initial states, \bob is not able to recover the ground truth as they require long range reasoning.
When the image is too ambiguous, local context may be insufficient to learn the correct dynamics.
Furthermore, despite the small network size, the NCAs training is more complex due to its iterative nature. It has to not only learn from the initial masks, but also learn how to correct and balance its own predictions into a steady state solution, which can result in slower training times than single pass methods.

A practical advantage of \bob is that it offers a drop-in mechanism that does not need to interact with the base model training process.
Its smaller parameter count (see Appendix \ref{app:liver-metrics}) and architecture makes it portable and ideal to be deployed in edge devices.
This \bob an attractive refinement layer for medical segmentation pipelines. Its steady-state solution preserves accurate segmentation while correcting those that are suboptimal.

\section{Conclusion \& Future Research}
\label{sec:conclusion}
This work introduces Neural Cellular Automata as a simple, data-driven mechanism for repairing segmentation masks.
By learning local repair dynamics from the data, \bob improves structural correctness across tasks with minimal integration effort.
The approach provides an attractive alternative to topology-aware training strategies and simplifies the refinement approach by using already explored techniques in the field of growth dynamics.
Looking forward, the main challenges lie in extending the method beyond proof-of-concept scenarios and into more integrated pipelines. 
An exciting future path is that of refining volumetric data, as well as designing more general refiners, like \bob, capable of handling multiple error types.
Addressing these would result in an off-the-shelf refinement method for a large domain of tasks.

Overall, this study shows that NCA offer a natural and surprisingly effective fit for segmentation refinement.
They are simple, easy to deploy, and capable of repairing errors that compromise downstream reliability without task-specific customization of the refiner, opening exciting new directions in refinement solutions for clinical settings.

\bibliographystyle{unsrtnat}
\bibliography{references}

\begin{thebibliography}{49}
\providecommand{\natexlab}[1]{#1}
\providecommand{\url}[1]{\texttt{#1}}
\expandafter\ifx\csname urlstyle\endcsname\relax
  \providecommand{\doi}[1]{doi: #1}\else
  \providecommand{\doi}{doi: \begingroup \urlstyle{rm}\Url}\fi

\bibitem[Long et~al.(2015)Long, Shelhamer, and Darrell]{Long_CVPR_2015_FCN}
Jonathan Long, Evan Shelhamer, and Trevor Darrell.
\newblock {Fully Convolutional Networks for Semantic Segmentation}.
\newblock In \emph{IEEE Conference on Computer Vision and Pattern Recognition
  (CVPR)}, pages 3431--3440, 2015.

\bibitem[Ronneberger et~al.(2015)Ronneberger, Fischer, and
  Brox]{Ronneberger_MICCAI_2015_UNet}
Olaf Ronneberger, Philipp Fischer, and Thomas Brox.
\newblock {U-Net: Convolutional Networks for Biomedical Image Segmentation}.
\newblock In \emph{Medical Image Computing and Computer-Assisted Intervention
  (MICCAI)}, volume 9351, pages 234--241, 2015.
\newblock \doi{10.1007/978-3-319-24574-4_28}.

\bibitem[Huang et~al.(2019)Huang, Wang, Huang, Huang, Wei, and
  Liu]{Huang_ICCV_2019_CCNet}
Zilong Huang, Xinggang Wang, Lichao Huang, Chang Huang, Yunchao Wei, and Wenyu
  Liu.
\newblock {CCNet: Criss-Cross Attention for Semantic Segmentation}.
\newblock In \emph{IEEE/CVF International Conference on Computer Vision
  (ICCV)}, pages 603--612, 2019.
\newblock \doi{10.1109/ICCV.2019.00069}.

\bibitem[Yuan et~al.(2020{\natexlab{a}})Yuan, Chen, and
  Wang]{Yuan_ECCV_2020_OCR}
Yuhui Yuan, Xilin Chen, and Jingdong Wang.
\newblock {Object-Contextual Representations for Semantic Segmentation}.
\newblock In \emph{European Conference on Computer Vision (ECCV)}, volume
  12351, pages 173--190, 2020{\natexlab{a}}.
\newblock \doi{10.1007/978-3-030-58539-6_11}.

\bibitem[Liu et~al.(2021)Liu, Lin, Cao, Hu, Wei, Zhang, Lin, and
  Guo]{Liu_ICCV_2021_Swin}
Ze~Liu, Yutong Lin, Yue Cao, Han Hu, Yixuan Wei, Zheng Zhang, Stephen Lin, and
  Baining Guo.
\newblock {Swin Transformer: Hierarchical Vision Transformer Using Shifted
  Windows}.
\newblock In \emph{IEEE/CVF International Conference on Computer Vision
  (ICCV)}, pages 10012--10022, 2021.
\newblock \doi{10.1109/ICCV48922.2021.00986}.

\bibitem[Wyburd et~al.(2024)Wyburd, Dinsdale, Jenkinson, and
  Namburete]{wyburd2024anatomically}
Madeleine~K Wyburd, Nicola~K Dinsdale, Mark Jenkinson, and Ana~IL Namburete.
\newblock {Anatomically plausible segmentations: Explicitly preserving topology
  through prior deformations}.
\newblock \emph{Medical Image Analysis}, 97:\penalty0 103222, 2024.

\bibitem[Clough et~al.(2020)Clough, Byrne, Oksuz, Zimmer, Schnabel, and
  King]{Clough_TPAMI_2020_TopoLossPH}
James~R. Clough, Nicholas Byrne, Ilkay Oksuz, Veronika~A. Zimmer, Julia~A.
  Schnabel, and Andrew~P. King.
\newblock {A Topological Loss Function for Deep-Learning Based Image
  Segmentation Using Persistent Homology}.
\newblock \emph{IEEE Transactions on Pattern Analysis and Machine
  Intelligence}, 2020.
\newblock \doi{10.1109/TPAMI.2020.3035294}.

\bibitem[Shit et~al.(2021)Shit, Paetzold, Sekuboyina, Ezhov, Unger, Zhylka,
  Pluim, Bauer, and Menze]{shit2021cldice}
Suprosanna Shit, Johannes~C Paetzold, Anjany Sekuboyina, Ivan Ezhov, Alexander
  Unger, Andrey Zhylka, Josien~PW Pluim, Ulrich Bauer, and Bjoern~H Menze.
\newblock {clDice-a novel topology-preserving loss function for tubular
  structure segmentation}.
\newblock In \emph{Proceedings of the IEEE/CVF conference on computer vision
  and pattern recognition}, pages 16560--16569, 2021.

\bibitem[Kervadec et~al.(2021)Kervadec, Bouchtiba, Desrosiers, Granger, Dolz,
  and Ayed]{Kervadec_MedIA_2021_BoundaryLoss}
Hoel Kervadec, Jihene Bouchtiba, Christian Desrosiers, Eric Granger, Jose Dolz,
  and Ismail~Ben Ayed.
\newblock {Boundary Loss for Highly Unbalanced Segmentation}.
\newblock \emph{Medical Image Analysis}, 67:\penalty0 101851, 2021.
\newblock \doi{10.1016/j.media.2020.101851}.

\bibitem[Wyburd et~al.(2021)Wyburd, Dinsdale, Namburete, and
  Jenkinson]{wyburd2021teds}
Madeleine~K Wyburd, Nicola~K Dinsdale, Ana~IL Namburete, and Mark Jenkinson.
\newblock {TEDS-Net: enforcing diffeomorphisms in spatial transformers to
  guarantee topology preservation in segmentations}.
\newblock In \emph{International Conference on Medical Image Computing and
  Computer-Assisted Intervention}, pages 250--260, 2021.

\bibitem[Alonso and Kirkegaard(2023)]{alonso2023fast}
Albert Alonso and Julius~B Kirkegaard.
\newblock {Fast detection of slender bodies in high density microscopy data}.
\newblock \emph{Communications Biology}, 6\penalty0 (1):\penalty0 754, 2023.

\bibitem[Kirchhoff et~al.(2024)Kirchhoff, Rokuss, Roy, Kovacs, Ulrich, Wald,
  Zenk, Vollmuth, Kleesiek, Isensee, et~al.]{kirchhoff2024skeleton}
Yannick Kirchhoff, Maximilian~R Rokuss, Saikat Roy, Balint Kovacs, Constantin
  Ulrich, Tassilo Wald, Maximilian Zenk, Philipp Vollmuth, Jens Kleesiek,
  Fabian Isensee, et~al.
\newblock {Skeleton recall loss for connectivity conserving and resource
  efficient segmentation of thin tubular structures}.
\newblock In \emph{European Conference on Computer Vision}, pages 218--234,
  2024.

\bibitem[Huang et~al.(2024)Huang, Li, Shen, and Xu]{huang2024meta}
Shiqi Huang, Jianan Li, Ning Shen, and Tingfa Xu.
\newblock {Meta-tubular-net: A robust topology-aware re-weighting network for
  retinal vessel segmentation}.
\newblock \emph{Biomedical Signal Processing and Control}, 91:\penalty0 106060,
  2024.

\bibitem[Song et~al.(2025)Song, Huang, Liu, Islam, Yang, Wang, Zheng, and
  Wang]{song2025optimized}
Dongning Song, Weijian Huang, Jiarun Liu, Md~Jahidul Islam, Hao Yang, Shuqiang
  Wang, Hairong Zheng, and Shanshan Wang.
\newblock {Optimized Vessel Segmentation: A Structure-Agnostic Approach with
  Small Vessel Enhancement and Morphological Correction}.
\newblock \emph{IEEE Transactions on Image Processing}, 2025.

\bibitem[Amiri et~al.(2024)Amiri, Karimzadeh, Vrtovec, Gudmann Steuble~Brandt,
  Thomsen, Brun~Andersen, Felix~M{\"u}ller, Bertil~Rodell, and
  Ibragimov]{amiri2024centerline}
Sepideh Amiri, Reza Karimzadeh, Toma{\v{z}} Vrtovec, Erik Gudmann
  Steuble~Brandt, Henrik~S Thomsen, Michael Brun~Andersen, Christoph
  Felix~M{\"u}ller, Anders Bertil~Rodell, and Bulat Ibragimov.
\newblock {Centerline-guided reinforcement learning model for pancreatic duct
  identifications}.
\newblock \emph{Journal of Medical Imaging}, 11\penalty0 (6):\penalty0
  064002--064002, 2024.

\bibitem[Boykov and Jolly(2001)]{Boykov_ICCV_2001_GraphCuts}
Yuri~Y. Boykov and Marie-Pierre Jolly.
\newblock {Interactive Graph Cuts for Optimal Boundary \& Region Segmentation
  of Objects in N-D Images}.
\newblock In \emph{IEEE International Conference on Computer Vision (ICCV)},
  pages 105--112, 2001.
\newblock \doi{10.1109/ICCV.2001.937505}.

\bibitem[Soille et~al.(1999)]{Soille_2003_MorphologyBook}
Pierre Soille et~al.
\newblock \emph{Morphological image analysis: principles and applications},
  volume~2.
\newblock Springer, 1999.

\bibitem[Zheng et~al.(2015)Zheng, Jayasumana, Romera-Paredes, Vineet, Su, Du,
  Huang, and Torr]{Zheng_ICCV_2015_CRFRNN}
Shuai Zheng, Sadeep Jayasumana, Bernardino Romera-Paredes, Vibhav Vineet,
  Zhizhong Su, Dalong Du, Chang Huang, and Philip H.~S. Torr.
\newblock {Conditional Random Fields as Recurrent Neural Networks}.
\newblock In \emph{IEEE International Conference on Computer Vision (ICCV)},
  pages 1529--1537, 2015.
\newblock \doi{10.1109/ICCV.2015.179}.

\bibitem[Zdyb et~al.(2025)Zdyb, Alonso, and Kirkegaard]{zdyb2025spline}
Frans Zdyb, Albert Alonso, and Julius~B Kirkegaard.
\newblock {Spline refinement with differentiable rendering}.
\newblock In \emph{International Conference on Medical Image Computing and
  Computer-Assisted Intervention}, pages 558--567, 2025.

\bibitem[Dima et~al.(2025)Dima, Shit, Qiu, Holland, Mueller, Musio, Yang,
  Menze, Braren, Makowski, et~al.]{dima2025parametric}
Alina~F Dima, Suprosanna Shit, Huaqi Qiu, Robbie Holland, Tamara~T Mueller,
  Fabio Musio, Kaiyuan Yang, Bjoern Menze, Rickmer Braren, Marcus~R Makowski,
  et~al.
\newblock {Parametric shape models for vessels learned from segmentations via
  differentiable voxelization}.
\newblock In \emph{International Workshop on Shape in Medical Imaging}, pages
  247--261, 2025.

\bibitem[Mordvintsev et~al.(2020)Mordvintsev, Randazzo, Niklasson, and
  Levin]{mordvintsev2020growing}
Alexander Mordvintsev, Ettore Randazzo, Eyvind Niklasson, and Michael Levin.
\newblock Growing neural cellular automata.
\newblock \emph{Distill}, 5\penalty0 (2):\penalty0 e23, 2020.

\bibitem[Sudhakaran et~al.(2022)Sudhakaran, Najarro, and
  Risi]{sudhakaran2022goal}
Shyam Sudhakaran, Elias Najarro, and Sebastian Risi.
\newblock Goal-guided neural cellular automata: Learning to control
  self-organising systems.
\newblock \emph{arXiv preprint arXiv:2205.06806}, 2022.

\bibitem[Kalkhof et~al.(2023)Kalkhof, Gonz{\'a}lez, and
  Mukhopadhyay]{kalkhof2023med}
John Kalkhof, Camila Gonz{\'a}lez, and Anirban Mukhopadhyay.
\newblock {Med-nca: Robust and lightweight segmentation with neural cellular
  automata}.
\newblock In \emph{International Conference on Information Processing in
  Medical Imaging}, pages 705--716, 2023.

\bibitem[Kalkhof and Mukhopadhyay(2023)]{kalkhof2023m3d}
John Kalkhof and Anirban Mukhopadhyay.
\newblock {M3d-nca: Robust 3d segmentation with built-in quality control}.
\newblock In \emph{International Conference on Medical Image Computing and
  Computer-Assisted Intervention}, pages 169--178, 2023.

\bibitem[Ranem et~al.(2025)Ranem, Kalkhof, and Mukhopadhyay]{ranem2025ncadapt}
Amin Ranem, John Kalkhof, and Anirban Mukhopadhyay.
\newblock {NCAdapt: Dynamic adaptation with domain-specific Neural Cellular
  Automata for continual hippocampus segmentation}.
\newblock In \emph{2025 IEEE/CVF Winter Conference on Applications of Computer
  Vision (WACV)}, pages 3834--3843, 2025.

\bibitem[Neumann and Burks(1966)]{automata}
Johann Neumann and Arthur~W Burks.
\newblock \emph{Theory of self-reproducing automata}.
\newblock University of Illinois Press Urbana, 1966.

\bibitem[Pajouheshgar et~al.(2024)Pajouheshgar, Xu, and
  S{\"u}sstrunk]{pajouheshgar2024noisenca}
Ehsan Pajouheshgar, Yitao Xu, and Sabine S{\"u}sstrunk.
\newblock Noisenca: Noisy seed improves spatio-temporal continuity of neural
  cellular automata.
\newblock In \emph{Artificial Life Conference Proceedings 36}, volume 2024,
  page~57. MIT Press One Rogers Street, Cambridge, MA 02142-1209, USA
  journals-info~…, 2024.

\bibitem[Pajouheshgar et~al.(2023)Pajouheshgar, Xu, Zhang, and
  S{\"u}sstrunk]{pajouheshgar_DyNCARealTimeDynamic_2023}
Ehsan Pajouheshgar, Yitao Xu, Tong Zhang, and Sabine S{\"u}sstrunk.
\newblock Dynca: Real-time dynamic texture synthesis using neural cellular
  automata.
\newblock In \emph{Proceedings of the IEEE/CVF conference on computer vision
  and pattern recognition}, pages 20742--20751, 2023.

\bibitem[Palm et~al.(2022)Palm, Gonz{\'a}lez-Duque, Sudhakaran, and
  Risi]{palm2022variational}
Rasmus~Berg Palm, Miguel Gonz{\'a}lez-Duque, Shyam Sudhakaran, and Sebastian
  Risi.
\newblock Variational neural cellular automata.
\newblock \emph{arXiv preprint arXiv:2201.12360}, 2022.

\bibitem[Faldor and Cully(2024)]{faldor2024cax}
Maxence Faldor and Antoine Cully.
\newblock {Cax: Cellular automata accelerated in jax}.
\newblock \emph{arXiv preprint arXiv:2410.02651}, 2024.

\bibitem[Kalkhof et~al.(2025)Kalkhof, K{\"u}hn, Frisch, and
  Mukhopadhyay]{kalkhof2025parameter}
John Kalkhof, Arlene K{\"u}hn, Yannik Frisch, and Anirban Mukhopadhyay.
\newblock {Parameter-efficient diffusion with neural cellular automata}.
\newblock \emph{npj Unconventional Computing}, 2\penalty0 (1):\penalty0 10,
  2025.

\bibitem[Ranem et~al.(2024)Ranem, Kalkhof, and Mukhopadhyay]{ranem2024nca}
Amin Ranem, John Kalkhof, and Anirban Mukhopadhyay.
\newblock {Nca-morph: Medical image registration with neural cellular
  automata}.
\newblock \emph{arXiv preprint arXiv:2410.22265}, 2024.

\bibitem[Kirillov et~al.(2020)Kirillov, Wu, He, and
  Girshick]{Kirillov_CVPR_2020_PointRend}
Alexander Kirillov, Yuxin Wu, Kaiming He, and Ross Girshick.
\newblock {PointRend: Image Segmentation as Rendering}.
\newblock In \emph{IEEE/CVF Conference on Computer Vision and Pattern
  Recognition (CVPR)}, pages 9799--9808, 2020.
\newblock \doi{10.1109/CVPR42600.2020.00982}.

\bibitem[Zhang et~al.(2021)Zhang, Lu, Tan, Li, Zhang, Li, and
  Hu]{zhang2021refinemask}
Gang Zhang, Xin Lu, Jingru Tan, Jianmin Li, Zhaoxiang Zhang, Quanquan Li, and
  Xiaolin Hu.
\newblock {Refinemask: Towards high-quality instance segmentation with
  fine-grained features}.
\newblock In \emph{Proceedings of the IEEE/CVF conference on computer vision
  and pattern recognition}, pages 6861--6869, 2021.

\bibitem[Wu et~al.(2024)Wu, Chen, Liu, Yue, and Zhuang]{wu2024deep}
Qian Wu, Yufei Chen, Wei Liu, Xiaodong Yue, and Xiahai Zhuang.
\newblock Deep closing: Enhancing topological connectivity in medical tubular
  segmentation.
\newblock \emph{IEEE Transactions on Medical Imaging}, 43\penalty0
  (11):\penalty0 3990--4003, 2024.

\bibitem[Tang et~al.(2021)Tang, Chen, Li, Li, Zhang, and Hu]{tang2021look}
Chufeng Tang, Hang Chen, Xiao Li, Jianmin Li, Zhaoxiang Zhang, and Xiaolin Hu.
\newblock {Look closer to segment better: Boundary patch refinement for
  instance segmentation}.
\newblock In \emph{Proceedings of the IEEE/CVF conference on computer vision
  and pattern recognition}, pages 13926--13935, 2021.

\bibitem[Yuan et~al.(2020{\natexlab{b}})Yuan, Xie, Chen, and
  Wang]{yuan2020segfix}
Yuhui Yuan, Jingyi Xie, Xilin Chen, and Jingdong Wang.
\newblock Segfix: Model-agnostic boundary refinement for segmentation.
\newblock In \emph{European conference on computer vision}, pages 489--506,
  2020{\natexlab{b}}.

\bibitem[Zhou et~al.(2020)Zhou, Price, Cohen, Wilensky, and
  Davis]{zhou2020deepstrip}
Peng Zhou, Brian Price, Scott Cohen, Gregg Wilensky, and Larry~S Davis.
\newblock {Deepstrip: High-resolution boundary refinement}.
\newblock In \emph{Proceedings of the IEEE/CVF conference on computer vision
  and pattern recognition}, pages 10558--10567, 2020.

\bibitem[Cheng et~al.(2020)Cheng, Chung, Tai, and Tang]{cheng2020cascadepsp}
Ho~Kei Cheng, Jihoon Chung, Yu-Wing Tai, and Chi-Keung Tang.
\newblock {Cascadepsp: Toward class-agnostic and very high-resolution
  segmentation via global and local refinement}.
\newblock In \emph{Proceedings of the IEEE/CVF conference on computer vision
  and pattern recognition}, pages 8890--8899, 2020.

\bibitem[Wang et~al.(2023)Wang, Ding, Liew, Liu, Zhao, and Wei]{SegRefiner}
Mengyu Wang, Henghui Ding, Jun~Hao Liew, Jiajun Liu, Yao Zhao, and Yunchao Wei.
\newblock {SegRefiner: Towards Model-Agnostic Segmentation Refinement with
  Discrete Diffusion Process}.
\newblock In \emph{NeurIPS}, 2023.

\bibitem[Lagergren et~al.(2020)Lagergren, Rutter, and
  Flores]{lagergren2020region}
John Lagergren, Erica Rutter, and Kevin Flores.
\newblock {Region growing with convolutional neural networks for biomedical
  image segmentation}.
\newblock \emph{arXiv preprint arXiv:2009.11717}, 2020.

\bibitem[Lagergren et~al.(2023)Lagergren, Pavicic, Chhetri, York, Hyatt,
  Kainer, Rutter, Flores, Bailey-Bale, Klein, et~al.]{lagergren2023few}
John Lagergren, Mirko Pavicic, Hari~B Chhetri, Larry~M York, Doug Hyatt, David
  Kainer, Erica~M Rutter, Kevin Flores, Jack Bailey-Bale, Marie Klein, et~al.
\newblock {Few-shot learning enables population-scale analysis of leaf traits
  in Populus trichocarpa}.
\newblock \emph{Plant Phenomics}, 5:\penalty0 0072, 2023.

\bibitem[Millson(1976)]{millson1976first}
John~J Millson.
\newblock On the first betti number of a constant negatively curved manifold.
\newblock \emph{Annals of Mathematics}, 104\penalty0 (2):\penalty0 235--247,
  1976.

\bibitem[Kavur et~al.(2019)Kavur, Selver, Dicle, Barış, and
  Gezer]{CHAOSdata2019}
Ali~Emre Kavur, M.~Alper Selver, Oğuz Dicle, Mustafa Barış, and N.~Sinem
  Gezer.
\newblock {CHAOS - Combined (CT-MR) Healthy Abdominal Organ Segmentation
  Challenge Data}, April 2019.
\newblock URL \url{https://doi.org/10.5281/zenodo.3362844}.

\bibitem[Staal et~al.(2004)Staal, Abr{\`a}moff, Niemeijer, Viergever, and
  Van~Ginneken]{staal2004ridge}
Joes Staal, Michael~D Abr{\`a}moff, Meindert Niemeijer, Max~A Viergever, and
  Bram Van~Ginneken.
\newblock {Ridge-based vessel segmentation in color images of the retina}.
\newblock \emph{IEEE transactions on medical imaging}, 23\penalty0
  (4):\penalty0 501--509, 2004.

\bibitem[Kr{\"a}henb{\"u}hl and Koltun(2011)]{Krahenbuhl_NIPS_2011_DenseCRF}
Philipp Kr{\"a}henb{\"u}hl and Vladlen Koltun.
\newblock {Efficient Inference in Fully Connected CRFs with Gaussian Edge
  Potentials}.
\newblock In \emph{Advances in Neural Information Processing Systems
  (NeurIPS)}, volume~24, 2011.

\bibitem[Bernard et~al.(2018)Bernard, Lalande, Zotti, Cervenansky, Yang, Heng,
  Cetin, Lekadir, Camara, Ballester, et~al.]{bernard2018deep}
Olivier Bernard, Alain Lalande, Clement Zotti, Frederick Cervenansky, Xin Yang,
  Pheng-Ann Heng, Irem Cetin, Karim Lekadir, Oscar Camara, Miguel
  Angel~Gonzalez Ballester, et~al.
\newblock Deep learning techniques for automatic mri cardiac multi-structures
  segmentation and diagnosis: is the problem solved?
\newblock \emph{IEEE transactions on medical imaging}, 37\penalty0
  (11):\penalty0 2514--2525, 2018.

\bibitem[Hu et~al.(2019)Hu, Li, Samaras, and Chen]{hu2019topology}
Xiaoling Hu, Fuxin Li, Dimitris Samaras, and Chao Chen.
\newblock {Topology-preserving deep image segmentation}.
\newblock \emph{Advances in neural information processing systems}, 32, 2019.

\bibitem[Cao et~al.(2022)Cao, Wang, Chen, Jiang, Zhang, Tian, and
  Wang]{cao2022swin}
Hu~Cao, Yueyue Wang, Joy Chen, Dongsheng Jiang, Xiaopeng Zhang, Qi~Tian, and
  Manning Wang.
\newblock {Swin-unet: Unet-like pure transformer for medical image
  segmentation}.
\newblock In \emph{European conference on computer vision}, pages 205--218,
  2022.

\end{thebibliography}

\newpage
\appendix
\section{Model Architecture}
\label{app:model}
The refiner follows a Neural Cellular Automaton formulation in which each pixel holds a $K$-dimensional state vector.
In \bob the first channel encodes the visible mask, while the remaining $K-1$ channels act as latent variables.
We use  $K=16$ unless otherwise stated.
During inference, the visible state is initialized from the predicted mask, and all latent channels start at zero.

Local information comes from two sources.
The current state $s_t \in \R^{H\times W \times K}$, where $H$ \& $W$ denotes the image size, is processed by a learnable perception operator $g_{\phi_s}(s_t): \R^{H\times W \times K} \rightarrow \R^{H\times W \times K}$. 
It applies a small set of learnable spatial filters, initialized as the original identity, Sobel (x and y). 
We chose to have all filters learnable to give the network complete freedom. 
In parallel, the image $x \in \R^{H\times W \times C_x}$ with $C_x$ denoting the number of image channels, is processed by a separate learnable convolutional layer $g_{\phi_x}(x): \R^{H\times W \times C_x} \rightarrow \R^{H\times W \times N}$ that extracts local appearance cues at the same spatial resolution as the mask.
We limit this layer to be a $3x3$ convolutional layer whose output matches the dimension of $g_{\phi_s}(s_t)$ output ($N{=}64$ for $K{=}16$). 
This way the update step receives equal contribution from the image and the state.
The outputs of these two components are concatenated and passed to a lightweight transition network, implemented as a two-layer MLP with ReLU activations and a hidden width of~128.
The update rule is additive, akin to an Euler integration step with $\Delta t = 1$:
\begin{equation}
    s_{t+1} = s_t + f_\theta\!\left(g_{\phi}(s_t, x)\right).
    \label{eq:update-rule-app}
\end{equation}

A pixel is considered \emph{alive} if its visible value or that of any of its 8-neighbors exceeds~0.1, before and after the update step. Formally, let \(\Omega \subset \mathbb{Z}^2\) be the set of pixel coordinates. The 8-neighbourhood of a pixel \(v = (i,j) \in \Omega\), including \(v\) itself, is defined as:

\[
N_8^+(v) = \{(m,n) \in \Omega \mid |m-i| \le 1, \; |n-j| \le 1 \}.
\]

Let \(\alpha = 0.1\). The masking rule for \(s_{t+1}\)(v), i.e. K-state vector of a pixel $v$ is:
\[
s_{t+1}(v) :=
\begin{cases}
s_{t+1}(v), & \text{if }
\displaystyle \max_{q \in N_8^+(v)} \hat{y}_{t+1}(q) \ge \alpha
\;\text{and}\;
\displaystyle \max_{q \in N_8^+(v)} \hat{y}_{t}(q) \ge \alpha, \\[8pt]
0, & \text{otherwise}.
\end{cases}
\]
Only alive pixels may update the visible channel, while all pixels update latent channels at every step.
To improve robustness and following the original work, each pixel independently skips its update with probability $\tau=0.5$, introducing controlled stochasticity into the dynamics.

\subsection{Training Process}
\label{app:training}
The refiner learns to reconstruct clean labels from inputs that contain realistic segmentation errors.
The source of these errors differs across tasks.
For datasets such as ACDC and the synthetic liver experiment, we generate corrupted masks directly from the ground-truth labels.
The reason behind this being either to have a more controlled topological repair study or because topological errors, despite being critical, are less prominent on the resulting predictions, and thus the refiner does not focus on those artifacts.
These corruptions are created using simple morphological perturbations, such as thinning, small deletions, or mild opening and closing operations, which emulate typical segmentation failures.

For the retinal vessel segmentation task, the situation is different.
Thin structures often break in practice, and UNet predictions reflect these characteristic failure modes.
To expose the refiner to such realistic errors, we train it directly on UNet predictions paired with the ground-truth labels.
Although the model is trained on UNet outputs, we also evaluate it on top of Swin predictions to show that the learned dynamics generalize across architectures.
The training procedure is identical regardless of how corrupted masks are generated.
We maintain a pool of 256 evolving states.
Each iteration draws a minibatch with a batch size of $B=32$ and randomly replaces $N=2$ with new initial states $s_0$ constructed from fresh corrupted masks $y_0$.
The automaton is unrolled for 64 steps under Eq.~\eqref{eq:update-rule-app}.
A late iteration is selected uniformly between steps~32 and~64, and the loss is computed as the squared difference between the visible channel and the corresponding ground-truth mask.
Optimization uses AdamW with a learning rate of $10^{-4}$ and decays $\alpha_1{=}0.9$ and $\alpha_2{=}{0.999}$.
No topology-specific losses or additional data augmentations are used.

\section{Experiments: Liver (CHAOS-CT)}\label{app:liver-metrics}
\paragraph{Dataset}
Each CT slice is cropped to $128\times128$ around the liver.
Corruptions remove or modify regions to create holes, missing patches, or eroded boundaries.
The dataset is split by patient into 14 training, 3 validation, and 3 test subjects.
Each patient contains $\sim10$ liver slices, which we augment and generate $3\times$ perturbations of initial states, each with randomly removed chunks as exemplified in \figureref{fig:fill}.
There removed sections are rectangles or arbitrary sizes as well as circles, which we allow to overlap to create more variate of holes.

\paragraph{Quantitative Results on Liver Filling Task}

For completeness, %
reports the full set of liver-filling results, including mean and standard deviation for all metrics.
These values complement the main text by showing the variance across test slices and the relative performance of the initial masks, morphological reconstruction, and \bob.

\begin{table}[h]
\centering
\caption{Liver filling task: full metrics with mean $\pm$ std.}
\begin{tabular}{lccc}
\toprule
Method & HD $\downarrow$ & ASSD $\downarrow$ & DICE $\uparrow$ \\
\midrule
Initial mask      & $10.51 \pm 3.68$ & $0.673 \pm 0.887$ & $0.798 \pm 0.145$ \\
Morph. reconstr.  & $7.57 \pm 4.28$  & $0.190 \pm 0.790$ & $0.958 \pm 0.090$ \\
\textbf{\bob}     & $3.66 \pm 3.20$  & $0.094 \pm 0.779$ & $0.983 \pm 0.089$ \\
\bottomrule
\end{tabular}
\end{table}

\section{Experiments: Retinal Vessels (DRIVE) }\label{app:drive}

\paragraph{Dataset and Training}
The DRIVE \cite{staal2004ridge} dataset consists of 20 images, developed to provide a standardized benchmark for evaluating retinal blood vessel segmentation methods in the context of diabetic retinopathy screening. 
Due to its small dataset size we randomly split the data 0.8:0.2 across three folds and report the average validation metrics, as done in prior work see \cite{hu2019topology}. 
From the full size image we extract patches of size 64x64 for the NCA training using a sliding window with 0.25 overlap. We disregard empty initial predictions as there is no signal to refine. We do this for each fold and each baseline model we try to refine. This creates \~4000 patches for all 20 images. 

\paragraph{Baselines}
The initial predictions are generated using a UNet and Swin. The UNet is configured with five encoder levels with channel sizes (16, 32, 64, 128, 256) and downsampling strides of 2 at each level. Each block contains 2 residual units with a 5×5 kernel, and the decoder uses 3×3 upsampling kernels. The Swin-UNETR we configure feature size 48, where the encoder uses Swin Transformer v2 blocks with patch size 2, depths 2 at each level, number of heads of 3, 6, 12, 24 respectively and a window size of 7. Trained with a Dice loss and AdamW with a learning rate of $10^{-2}$

Each model is trained to predict patches of size 128x128, randomly sampled during training. 
We also apply random flips, rotation, zoom, and random grayscale blending to the samples before cropping. 
During inference, we predict the full size image using a sliding window with 0.5 overlap and Gaussian averaging of the predicted logits across the patches. 
The model with the highest Dice score on the full sized validation images is selected for each fold and used to create the initial predictions that we will refine. 

Furthermore, we use UNet and test its ability to refine the initial prediction and turn the prediction problem into a residual learning task. Now, trained on the same 64x64 patches as \bob, the UNet receives the concatenation of initial prediction and image. This is to see how well the UNet compares in the same setting. 

We compare \bob to three popular segmentation refinement methods: SegFix~\cite{yuan2020segfix},SegRefiner~\cite{SegRefiner} and DenserCRF~\cite{Krahenbuhl_NIPS_2011_DenseCRF}. 
Each refiner was trained on the same dataset as \bob, until convergence, using their publicly available implementations for Segfix\footnote{\url{https://github.com/openseg-group/openseg.pytorch}}, SegRefiner\footnote{\url{https://github.com/MengyuWang826/SegRefiner}} and DenserCRF \footnote{\url{https://github.com/heiwang1997/DenseCRF}}.

\begin{figure}[tb]
\begin{minipage}[c]{0.6\linewidth}
  \centering
  {\includegraphics[height=4cm]{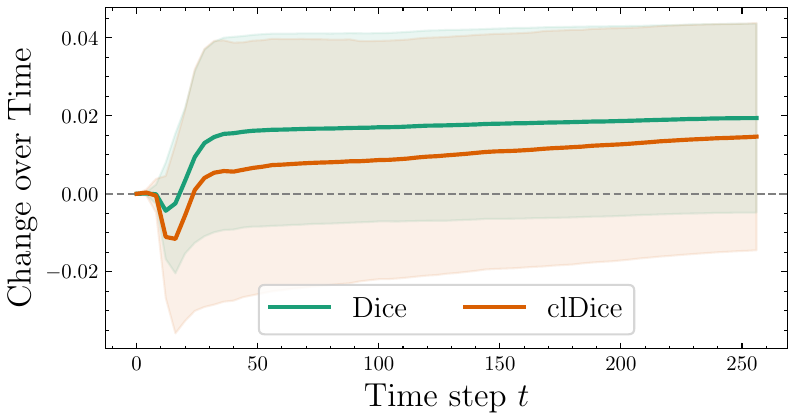}}
\end{minipage}
\begin{minipage}[c]{.375\linewidth}
 \centering
    \includegraphics[height=4cm]{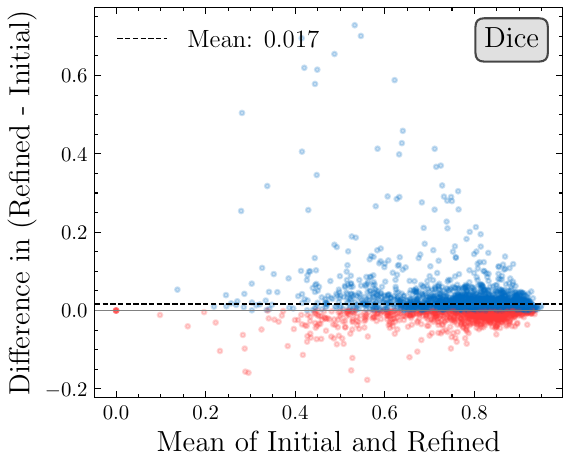}
\end{minipage}
\caption{\textbf{Retina vessel refinement.} (left) clDice and Dice progression during the refinement steps. Here we display the mean and std over all images. (right) Displays a Bland-Altman Plot comparing initial and refined Dice scores of each patch.}
\label{fig:bridge-appendix}
\end{figure}

\paragraph{Additional Results}
Complementary to the results shown in Section \ref{subsec:exp-bridge} we add some additional results for completeness. \figureref{fig:bridge-appendix} (left) shows the same convergence behaviour for clDice and Dice during inference. Note there is a small drop in performance in the early states, which we attribute to an initial exploration phase, as no loss is computed for $T<32$. Furthermore, \figureref{fig:bridge-appendix} (right) shows that the Dice scores follow the same pattern observed for clDice, supporting the consistency of \bob’s behaviour at the patch level.

\paragraph{Ablation Experiments}
\label{app:abblation}

To better understand which components drive the refinement behavior, we perform an ablation study on this task.
We systematically vary the number of state channels $K$, the pool-replacement rate $N$, and the number of unrolled steps $T$, and additionally compare variants with and without an image encoder, as well as supervision from perturbed labels versus model predictions.
All variants are trained with the same protocol and evaluated on DICE, clDICE, and Betti number deviations.
As summarized in \tableref{tab:ablation}, this setup isolates the effect of architectural capacity, state refresh frequency, rollout depth, and supervision source on the ability of \bob to correct fragmented vessels. Here we trained the model to repair vessels perturbed by morphological operations. Each label randomly gets 75\% of their thin structures removed by erosion, and returns the mask with most thin structures removed. This allows for a more controlled environment to analyse the models behavior. Ablation is done on a random fold, different to the training folds. We can see by decreasing the rollout depth too much (e.g. $T=16$) or replace the whole batch $N=B$, the model does not achieve a steady state. Both configuration limit (or prevent completely) the progression of samples and hence lead to instability during inference. 
Furthermore, limiting model capacity $K=1$ decreases performance, which tells us that the hidden channels learn important information about the growth directions. Interestingly, increasing the hidden channels further did not contribute any performance gains, presumably due to redundant information saturating the system. Increasing roll-out time ($T=128$) furthermore did not seem to improve the performance much as \bob already reaches a steady state solution with less memory cost. Lastly, we tried a formulation of \bob (Static) close to the original NCA, in which both image channels and states are passed through pre-defined non-learnable filters and concatenated afterwards. This limited performance significantly, most likely due to limited information from the image influencing the update step.  

\paragraph{Runtime analysis}

A key requirement for a segmentation model is that it is able to perfrom well on edge devices. 
Here we should that \bob runtime on refining 64x64 patches compared to the other refiners, as well as its parameter count, Table \ref{app:runtirme}.
Because refinement is iterative, \bob uses 64 update steps per image (the repair phase shown in \figureref{fig:bridge})
Despite this, its end-to-end runtime is on par with or faster than most learned refiners, while its parameter count remains extremely small ($12{,}048$), comparable to SegRefiner and orders of magnitude lighter than UNet- or SegFix-based refinement. This makes \bob one of the most computationally efficient refinement mechanisms among the evaluated methods.

\label{app:runtirme}
\begin{table}[htb]
\centering
\caption{Runtime and parameter comparison for CPU and GPU execution.}
\label{tab:runtime}
\begin{tabular}{lccc}
\toprule
\multicolumn{4}{c}{\textbf{CPU Results}} \\
\midrule
\textbf{Method} & \textbf{Time (s)} & \textbf{Time/Image (ms)} & \textbf{Parameters} \\
\midrule
DenseCRF      & $0.0195 \pm 0.0001$ & 3.90                 & N/A \\
SegFix        & $0.0246 \pm 0.0013$ & 4.91                 & 1,931,392 \\
SegRefiner    & $0.0023 \pm 0.0000$ & 0.46                 & 9,601 \\
UNet          & $0.0213 \pm 0.0000$ & 4.27                 & 3,976,476 \\
\bob          & $0.376 \pm 0.177$ & 75.20 (1.17\,ms/iter) & 12,048 \\
\midrule
\multicolumn{4}{c}{\textbf{GPU Results}} \\
\midrule
\textbf{Method} & \textbf{Time (s)} & \textbf{Time/Image (ms)} & \textbf{Parameters} \\
\midrule
SegFix        & $0.0133 \pm 0.0164$ & 2.66                 & 1,931,392 \\
SegRefiner    & $0.0011 \pm 0.0010$ & 0.23                 & 9,601 \\
UNet          & $0.0102 \pm 0.0001$ & 2.05                 & 3,976,476 \\
\bob          & $0.0124 \pm 0.0001$ & 2.48 (0.039\,ms/iter) & 12,048 \\
\bottomrule
\end{tabular}
\end{table}

\begin{table}[htb]
\label{tab:ablation}
\caption{
Results from the ablation study assessing how architectural and training choices affect the refiner. 
The table compares different state channel sizes $K$, pool–replacement rates $N$, and loss timestep $T$, 
along with the impact of adding/removing a learnable image encoder and training on perturbed labels instead of predictions. \bob is initialized with $(K=16, N=2, T=64)$
The comparison highlights which configurations yield more reliable repair of thin structures.
}
\begin{tabular}{lcccc}
\toprule
               & \textbf{DICE} $\uparrow$ & \textbf{clDICE} $\uparrow$ & $\Delta \beta_0$ $\downarrow$ & $\Delta \beta_1$ $\downarrow$ \\ \midrule
 (Perturbed Labels)            & 0.8518 & 0.7919 & 1.1801  & 2.0233  \\
\bob                     & 0.9088 & 0.8799 & 0.2306  & 0.9197  \\ \midrule %
\bob ($K=1$)             & 0.8498 & 0.7914 & 1.0984  & 2.0622  \\
\bob ($K=32$)            & 0.8930 & 0.8513 & 0.5052  & 1.4378  \\ \midrule
\bob ($N=B/2$)           & 0.8971 & 0.8573 & 1.0531  & 1.3951  \\
\bob ($N=B$)             & 0.2073 & 0.2353 & 0.3653  & 109.044 \\ \midrule
\bob ($T=16$)            & 0.6113 & 0.5497 & 47.3303 & 3.6218  \\
\bob ($T=128$)           & 0.9109 & 0.8812 & 0.4573  & 1.1049  \\ \midrule
\bob (Static)          & 0.8879 & 0.8327 & 3.8394  & 1.3135  \\

\end{tabular}
\end{table}

\section{Experiments: Myocardium }\label{app:hearts}

\paragraph{Dataset and Training}
For myocardium segmentation, we used the ACDC dataset \cite{bernard2018deep}. This dataset contains two labelled 3D cardiac scans from 100 patients. From each scan, we extracted 2 myocardium-containing slices, cropped to 144 by 208 pixels. The dataset was split by patient into  85 (850 slices) training,  15 (150 slices) validation, and 10 (100 slices) for testing. 

To expose the refiner to possible topological artifacts, we generated initial states by performing multiple morphological augmentations on the ground truth.
Erosion and dilation were used to thin or thicken boundaries, sometimes breaking the topology. This is critical for ACDC dataset where different thickness may be expected at different arc segments and we require the refiner to be able to handle both types of operations.
Additional, binary opening was applied to promote the cases of ring breaking.
Lastly, as the initial predictions can be good and we do not want to over refine good initial states, we included unperturbed ground truth masks.
These targeted distortions created a consistent spectrum of defects, gaps, oversegmented regions, undersegmented regions, and isolated artifacts that the refinement model was explicitly trained to repair.

\paragraph{Baselines}

For the base segmentation models, we used the published predictions from transformer based Swin-UNETR~\cite{cao2022swin} and UNet~\cite{Ronneberger_MICCAI_2015_UNet}.
For topology-based comparisons, we likewise used the published predictions from two additional methods. The first is TopoLoss~\cite{hu2019topology}, a topological loss function that aligns the persistent homology of predicted probability maps with that of the ground-truth labels. The second is TEDS-Net~\cite{wyburd2021teds}, a spatial transformer network that learns a topology-preserving deformation field and warps a prior shape of known topology to each input. TEDS-Net requires explicit prior knowledge of the anatomical topology.
All model implementations and predictions follow the descriptions provided in~\cite{wyburd2024anatomically}.

\begin{figure}
    \centering
    \includegraphics[width=1\linewidth]{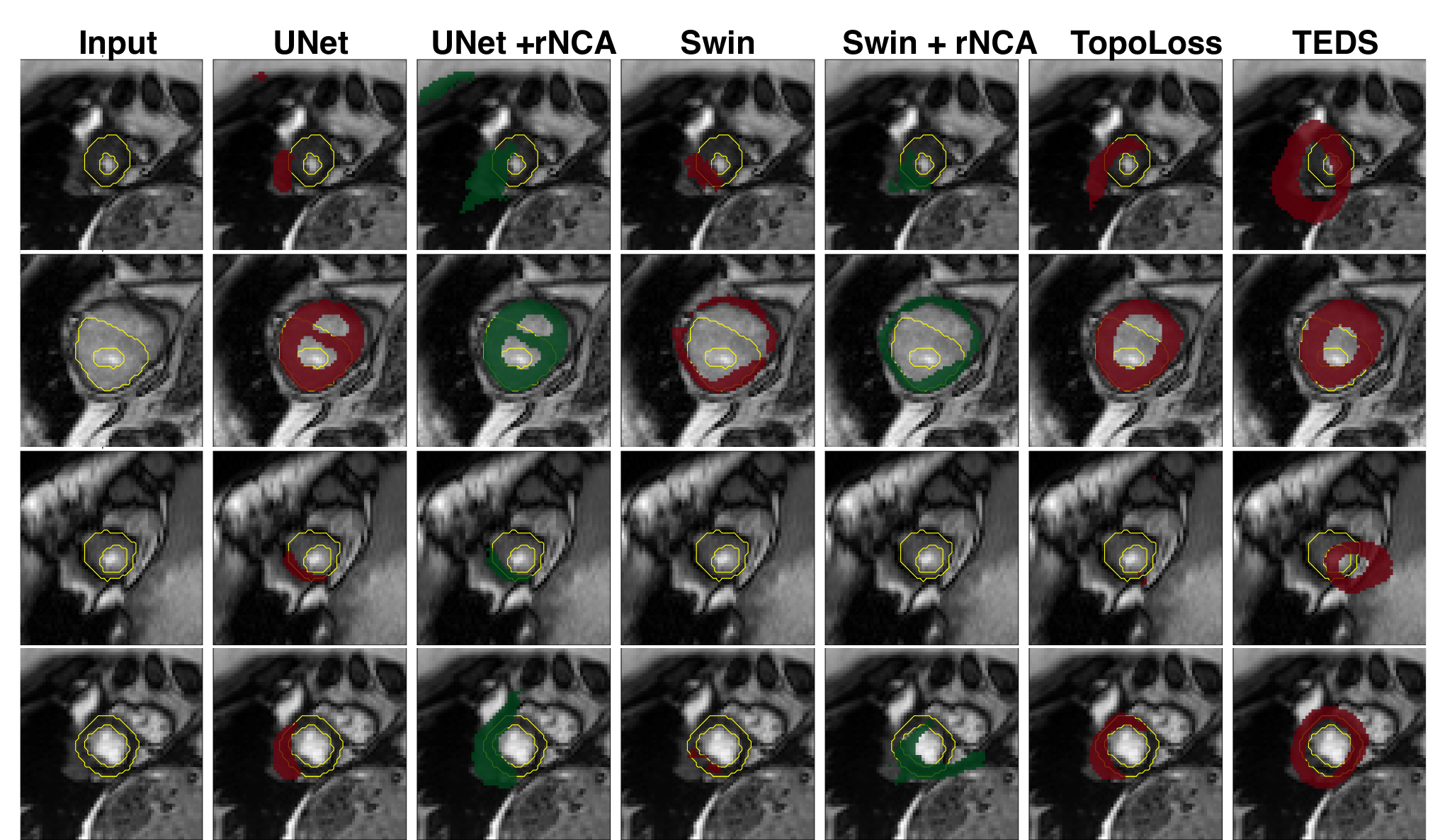}
    \caption{Examples of initially poor masks which \bob is unable to correct.}
    \label{fig:bad_hearts}
\end{figure}

\paragraph{Additional Results}
Figure \ref{fig:bad_hearts} shows examples of poor initial predictions from the base segmentation models that \bob is unable to correct. Further, it shows that despite TEDS-Net achieving 100\% correct topology, often the segmentations are misaligned. 
\end{document}